\documentclass[10pt,twocolumn,letterpaper]{article}

\usepackage{iccv}
\usepackage{times}
\usepackage{epsfig}
\usepackage{graphicx}
\usepackage{amsmath}
\usepackage{amssymb}
\usepackage[accsupp]{axessibility}  % Improves PDF readability for those with disabilities.

\usepackage{acronym}
\usepackage{multirow}
\usepackage{booktabs}
\usepackage{caption}
\usepackage{subcaption}

\acrodef{ac:DL}[DL]{\emph{Deep learning}}
\acrodef{ac:ML}[ML]{\emph{Machine learning}}
\acrodef{ac:SSL}[SSL]{\emph{Self-supervised learning}}
\acrodef{ac:NLP}[NLP]{\emph{Natural language processing}}
\acrodef{ac:CV}[CV]{\emph{Computer vision}}
\acrodef{ac:ViT}[ViT]{\emph{Visual Transformer}}
\acrodef{ac:ALAN}[ALAN]{\emph{Adaptive Locked Agnostic Network}}
\acrodef{ac:DINO}[DINO]{\emph{self-DIstillation with NO labels}}
\acrodef{ac:STEGO}[STEGO]{\emph{Self-supervised Transformer with Energy-based Graph Optimization}}
\acrodef{ac:DNN}[DNN]{\emph{Deep Neural Network}}
\acrodef{ac:ED}[ED]{\emph{end-diastole}}
\acrodef{ac:ES}[ES]{\emph{end-systole}}
\acrodef{ac:LV}[LV]{\emph{left ventricle}}
\acrodef{ac:LA}[LA]{\emph{left atrium}}
\acrodef{ac:A2C}[A2C]{\emph{apical 2-chamber}}
\acrodef{ac:A4C}[A4C]{\emph{apical 4-chamber}}
\acrodef{ac:CAMUS}[CAMUS]{\emph{Cardiac Acquisitions for Multi-structure Ultrasound Segmentation}}
\acrodef{ac:CRF}[CRF]{\emph{Conditional Random Field}}
\acrodef{ac:GELU}[GELU]{\emph{Gaussian Error Linear Unit}}
\acrodef{ac:RELU}[ReLU]{\emph{Rectified Linear Unit}}
\acrodef{ac:RAPTOR}[RAPTOR]{\emph{Robust And Parcelized Training Of echo Recordings}}
\acrodef{ac:STRAPTOR}[STRAPTOR]{\emph{Sequence-based Transformer for Robust And Parcelized Training Of echo Recordings}}

\usepackage[breaklinks=true,bookmarks=false]{hyperref}

\iccvfinalcopy % *** Uncomment this line for the final submission

\def\iccvPaperID{****} % *** Enter the ICCV Paper ID here

\ificcvfinal\fi

\begin{document}

%%%%%%%%% TITLE
\title{Unlocking the Heart Using Adaptive Locked Agnostic Networks}

\author{Sylwia Majchrowska\\
DSL\&D, BioPharma R\&D, AstraZeneca, \\
AI Sweden, \\
Gothenburg, Sweden \\
{\tt\small sylwia.majchrowska@astrazeneca.com}
\and
Anders Hildeman\\
DS\&AA, BioPharma R\&D, AstraZeneca, \\
Gothenburg, Sweden \\
{\tt\small anders.hildeman@astrazeneca.com}
\and
Philip Teare\\
CAI, BioPharma R\&D, AstraZeneca, \\
Cambridge, UK \\
{\tt\small philip.teare@astrazeneca.com}
\and
Tom Diethe\\
CAI, BioPharma R\&D, AstraZeneca, \\
Cambridge, UK \\
{\tt\small tom.diethe@astrazeneca.com}
}

\maketitle
\ificcvfinal\thispagestyle{empty}\fi

%%%%%%%%% ABSTRACT
\begin{abstract}
   Supervised training of deep learning models for medical imaging applications requires a significant amount of labeled data. This is posing a challenge as the images are required to be annotated by medical professionals. To address this limitation, we introduce the Adaptive Locked Agnostic Network (ALAN), a concept involving self-supervised visual feature extraction using a large \emph{backbone} model to produce anatomically robust semantic self-segmentation. 
   In the ALAN methodology, this self-supervised training occurs only once on a large and diverse dataset. Due to the intuitive interpretability of the segmentation, \emph{downstream} models tailored for specific tasks can be easily designed using \emph{white-box} models with few parameters. This, in turn, opens up the possibility of communicating the inner workings of a model with domain experts and introducing prior knowledge into it. It also means that the \emph{downstream} models become less data-hungry compared to fully supervised approaches. These characteristics make ALAN particularly well-suited for resource-scarce scenarios, such as costly clinical trials and rare diseases. 
   In this paper, we apply the ALAN approach to three publicly available echocardiography datasets: EchoNet-Dynamic, CAMUS, and TMED-2. Our findings demonstrate that the self-supervised \emph{backbone} model robustly identifies anatomical subregions of the heart in an apical four-chamber view. Building upon this, we design two \emph{downstream} models, one for segmenting a target anatomical region, and a second for echocardiogram view classification.
\end{abstract}

%%%%%%%%% BODY TEXT
\begin{table*}[!ht]
\centering
\caption{DICE scores of recent works regarding cardiac image segmentation of \emph{left ventricle}~(LV) and \emph{left atrium}~(LA) performed on the EchoNet-Dynamic~\cite{lit:EchoNetDynamic} and CAMUS~\cite{lit:CAMUS} datasets.}
\label{tab:sota_seg}
\resizebox{\textwidth}{!}{%
\begin{tabular}{ccccccc}
\toprule
 \multirow{3}{*}{\textbf{Source}} & \multirow{3}{*}{\textbf{Model}} & \multirow{3}{*}{\textbf{Training strategy}} & \multirow{3}{*}{\textbf{Training data}}  & \multicolumn{3}{c}{\textbf{Test data}} \\ 
    &   &   &   & \multicolumn{2}{c}{\textbf{CAMUS}} & \multicolumn{1}{c}{\textbf{EchoNet-Dynamic}} \\
    &   &   &   & \textbf{LV {[}\%{]}} &  \textbf{LA {[}\%{]}} & \textbf{LV {[}\%{]}}  \\ \midrule 
Stough \textit{et al.}, 2020~\cite{lit:Stough2020LeftVA}& UNet & fully-supervised & CAMUS & 0.93 & 0.89 & - \\
Wei \textit{et al.}, 2020~\cite{lit:Wei2020}  & 3D UNet & semi-supervised & CAMUS & 0.91 & 0.89 & - \\
Chen \textit{et al.}, 2021~\cite{lit:Chen2021} & 2.5D UNet & semi-supervised & CAMUS & 0.93 & 0.88 & - \\
Ouyang\textit{et al.}, 2021~\cite{lit:EchoNetDynamic} & DeepLabV3 & semi-supervised & EchoNet-Dynamic & - & - & 0.92 \\
Saeed \textit{et al.}, 2022~\cite{lit:saeed2022} & SimCLR + DeepLabV3 & self-supervised & EchoNet-Dynamic & 0.91 & - & 0.91 \\
\bottomrule  
\end{tabular}%
}
\end{table*}
\section{Introduction}

Examination of heart function is in clinical practise often conducted by analyzing cardiac ultrasound image sequences that visualize the beating heart of a patient. Traditionally, automating this task demand training a black-box model in a fully-supervised manner. This requires a large labeled dataset due to the high dimensionality of image sequences.
While it might be feasible to acquire a sufficient number of cardiac ultrasound sequences depicting both general healthy behavior and common pathologies, obtaining relevant medical annotations can be challenging, since it necessitates manual labeling by trained professionals.
Furthermore, the inherent subjectivity and variability associated with manual annotation, particularly for segmentation masks, can further complicate the learning of robust patterns between input and output. Moreover, in the case of rare diseases, acquiring a sufficient number of sequences to train a large neural network can be particularly challenging, even if labels are available.

To address these issues, the approach of \ac{ac:SSL} has recently grown in popularity~\cite{lit:DINO,lit:SimCLR,lit:byol}. \ac{ac:SSL} involves learning latent representation of the data completely without relying on labels. 
The basic idea of \ac{ac:SSL} is to automatically engineer features that are rich in informational content and can be used to distinguish structural differences in the data~\cite{lit:GoodBengCour16}. 
A smaller \emph{downstream} model can then be applied on top of these latent features, in order to solve the specific task at hand, typically trained in a supervised manner. 

% our contribiution
The main contributions of this work are: firstly, we introduce the \ac{ac:ALAN}-approach to echocardiogram data, solving multiple clinically relevant tasks using simple models on top of the same pretrained \ac{ac:DNN}; secondly, we apply the concept of parcelization (sub-segmenting) in self-supervised segmentation, demonstrating that target segments are typically derivatives of multiple parcels; finally, we show that the pretrained \ac{ac:DNN} does not necessarily need to be trained on echocardiograms itself, as latent features can generalize from other imaging modalities.

\section{Background and related works}

\ac{ac:DNN}s have shown remarkable success in computer vision tasks when trained on large datasets in fully supervised setting~\cite{lit:vit,lit:Hugo2021}. However, annotating data at the pixel level for segmentation is extremely laborious and time-consuming. \ac{ac:SSL} methods that leverage unlabelled data can help overcome this challenge. Several recent papers have explored using self-supervision to pretrain \ac{ac:DNN}s for image segmentation~\cite{lit:chaitanya2020contrastive} and classification~\cite{lit:Azizi9710396,lit:ZHANG2021491}. 
Despite the myriad of self-supervised techniques that have emerged in recent years~\cite{lit:Li23,lit:Zhu2020Rubiks}, for the purpose of this work, we focus on solutions adapted to echocardiography data.

Previous studies on echocardiograms using \ac{ac:SSL} methods have shown that various approaches can successfully be applied to tasks such as cardiac view classification~\cite{lit:anand2022,lit:chartsias2021} and segmentation~\cite{lit:saeed2022} of a specific cardiac region, as summarized in Table~\ref{tab:sota_seg}).

Chartsias \textit{et al.}~\cite{lit:chartsias2021} applied contrastive learning for the classification of 13 different cardiac views, achieving an F1 score of 89.2\% using data from both private and public sources. 
In their pipeline the fully-supervised head (the \emph{downstream} task) was trained on top of the self-supervised backbone (the \emph{backend} task) consisting of an encoder and projection network using a contrastive loss. Compared to a fully-supervised baseline model, they reported an improvement in terms of F1 score of up to 26\% for minimal labelled views. 

In the work of Anand \textit{et al.}~\cite{lit:anand2022}, the performance of both contrastive and non-contrastive \ac{ac:SSL} approaches for classifying cardiac views was tested. They compared performance when pretraining the \ac{ac:SSL} methods on ImageNet. In their setup, which included 8 standard cardiac views, classification with DINO ViT achieved an accuracy of 64\%. This revealed that \ac{ac:SSL} methods can learn data representations that can be fine-tuned for downstream cardiac view detection using only a few annotations.

A self-supervised pretraining approach for \ac{ac:LV} segmentation was proposed by Saeed \textit{et al.}~\cite{lit:saeed2022}. They showed that they could get DICE scores around 90\% for a supervised algorithm (DeepLabV3~\cite{lit:ChenPSA17} and UNet~\cite{lit:Ronneberger2015}) using a much smaller amount of training data if they pretrained the network using self-supervision (SimCLR~\cite{lit:SimCLR} and BYOL~\cite{lit:byol}). 

The previous work focused on pretraining a model for later fine-tuning, mainly in order to reduce the need for big datasets. That is, we do not propose retraining the last layers of a network with new data, nor do we suggest connecting another network on top of it. 
In our work we take a more holistic approach, focusing on learning a \emph{backbone}-model in order to lock it in for the future. This pretrained model can then be the foundation on which multiple and vastly different medical tasks can be solved using relatively simple and intuitive models, i.e., not using neural networks, tree-based models or large LASSO regressions for instance. 
In this work we present two examples of applying the \ac{ac:ALAN}-approach, a classification and a segmentation task based on echocardiogram sequences.

\begin{figure*}[!ht]
\begin{center}
\includegraphics[width=\textwidth]{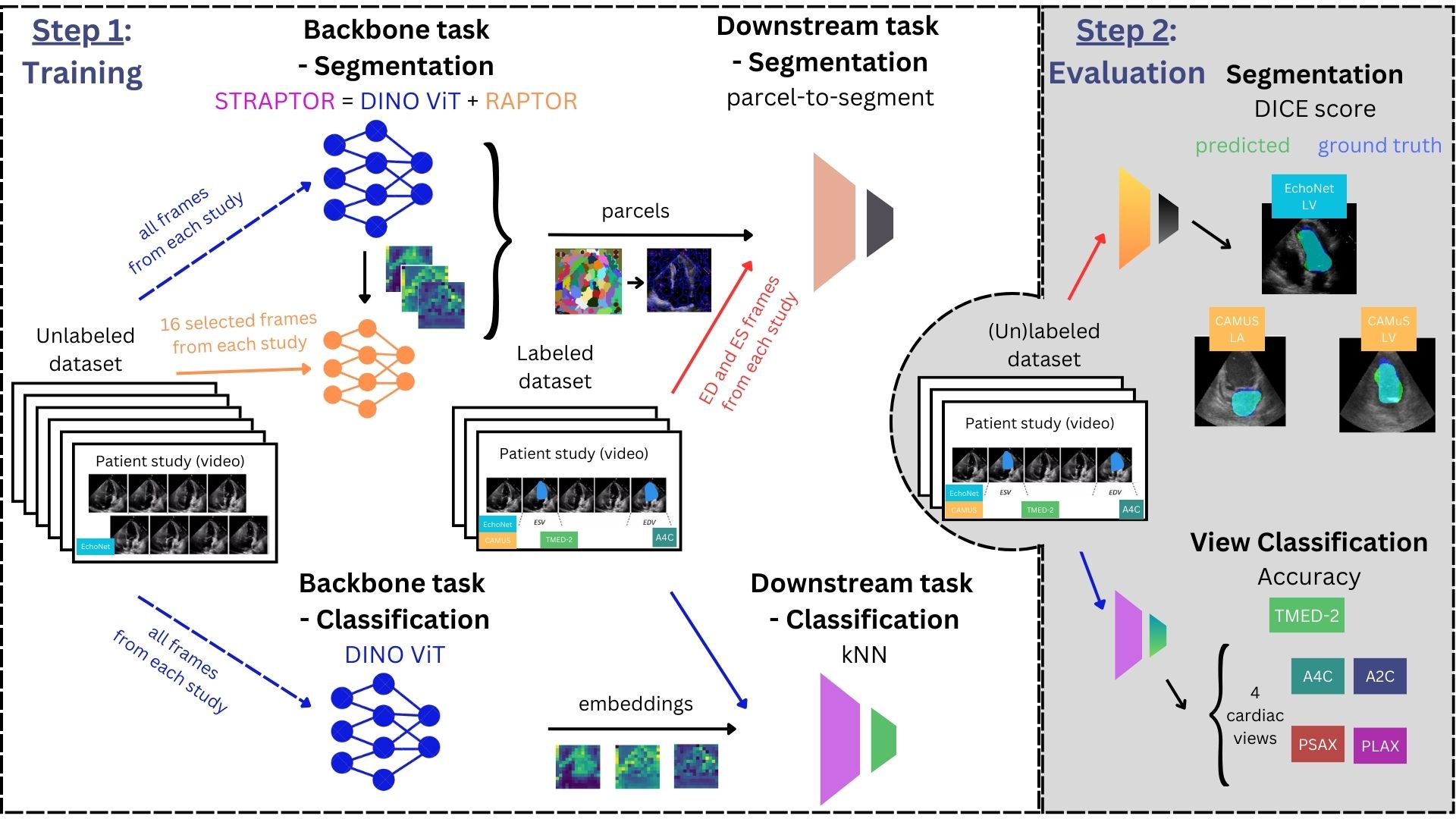}
\end{center}
   \caption{A sketch of our \ac{ac:ALAN}-approach for echocardiogram image segmentation and view-classification. For both downstream tasks we use DINO as the \emph{backbone}, which is extended by the RAPTOR for segmentation. In the training step we have 2 scenarios: DINO trained on echodiagram images, and DINO pretrained on ImageNet. In the evaluation step we calculated DICE scores for segmentation of two cardiac regions, and accuracy for view-classification. In each stage, disjoint subsets of different datasets~\cite{lit:huangTMED2Dataset2022,lit:CAMUS,lit:EchoNetDynamic} were used.}
\label{fig:pipeline}
\end{figure*}

\section{Methodology}
\label{sec:methods}
We propose the \ac{ac:ALAN}-approach, based on a pipeline consisting of a \emph{backbone} and \emph{downstream model}, as presented in Fig.~\ref{fig:pipeline}. The \emph{backbone} is a \ac{ac:DNN} trained on a large but more general dataset in a self-supervised manner. The \emph{downstream models} uses the same \emph{latent features} of the \emph{backbone} as input.

As for the \emph{backbone}, we use a \ac{ac:ViT}~\cite{lit:vit} architecture trained with \ac{ac:DINO}~\cite{lit:DINO}, a non-contrastive \ac{ac:SSL} method based on a teacher-student training mechanism.
For the view-classification task, the output of the \emph{backbone} model is directly fed into a, \emph{downstream}, weighted k-nearest neighbor (kNN) classifier~\cite{lit:wu2018unsupervised}.
For the segmentation task, the \emph{latent features} extracted from the DINO are projected through a self-supervised \emph{segmentation head} using our \ac{ac:RAPTOR} architecture based on the \ac{ac:STEGO} model~\cite{lit:stego}. In the context of the ALAN-approach for echocardiography videos segmentation, we consider the \ac{ac:RAPTOR} model to be part of the backbone and not a downstream task, and as a whole, the model is referred to as \ac{ac:STRAPTOR}. The reason for this is that it is learning by self-supervision and can also later be locked.

The \ac{ac:STEGO}~\cite{lit:stego} framework extends the \ac{ac:DINO}~\cite{lit:DINO} approach to achieve a completely self-supervised image segmentation. Their method acts as a plugin on top of the \ac{ac:DINO} output, distilling it into a segmented image with a maximum of $K$ number of classes. \ac{ac:STEGO} acts partly as a contrastive learner, comparing the signal from paired augmentations of the same image (attractive force) with augmentations of different images (repulsive force). In the original \ac{ac:STEGO}-algorithm, authors introduced a third class of image comparisons, using a kNN-classifier on the \ac{ac:DINO} output to identify images that are \textit{similar} to a target image. These similar images are then considered as examples of image pairs situated between the attractive and repulsive forces. 
However, since the echocardiogram dataset consists of video sequences, our setup differs from the still images used in the original work~\cite{lit:stego}. Therefore, we made some modifications of the original \ac{ac:STEGO} model which we refer to as \ac{ac:STRAPTOR}.

In \ac{ac:STRAPTOR}, we do not need a kNN-classifier to find \textit{similar} image pairs. Instead, we use different frames from the same sequence as \textit{similar} images, the same frame from the same sequence as the \textit{attractive force}, and random frames from different sequences as a \textit{repulsive force}. Our rationale for enforcing a comparison between frames of the same sequence is to encourage the discovery of segments that are robust across all phases of the cardiac cycle, potentially leading to segmentations that are stable over time. 

The \ac{ac:RAPTOR} architecture is implemented using a 3-layer perceptron with skip connections in the first and second layers. The first layer utilizes a \ac{ac:GELU} activation function~\cite{lit:hendrycks2023gaussian}, allowing for a slightly less linear relationship if needed, while the second layer uses a \ac{ac:RELU} activation function~\cite{lit:Nair2010RectifiedLU} which has a lower computational footprint than \ac{ac:GELU}. Both the first and second layer has equally many input dimensions as output dimensions. This dimensionality is equal to the number of channels of the DINO ViT architecture used as input (in this case ViT-S with a 384 channel output). 
The third layer has an output dimensionality of $K$ (the number of segmentation classes chosen) and a softmax activation function. 
The architecture's implementation is based on publicly accessible sources and is publicly available\footnote{https://github.com/AstraZeneca/UnlockingHeart}. 

STEGO projection head and \ac{ac:RAPTOR} are acting on a pixel-by-pixel basis on top of the DINO output. Since DINO is downsampling the image by a convolutional \textit{patching}-layer, the output of \ac{ac:RAPTOR} will have a lower resolution than the original input image. This is no issue during the training of \ac{ac:STRAPTOR} since it only compares the output of DINO with the output of \ac{ac:RAPTOR}. However, when using \ac{ac:RAPTOR} for self-segmentation one typically want it to have the same resolution as the original image. To handle this we are upsampling the \ac{ac:STRAPTOR} output by bi-linear interpolation just before the softmax activation function in the final layer of the architecture. The softmax activation is then applied on top of the high-resolution image instead. 

In comparison with the original STEGO we did not utilize a \ac{ac:CRF}~\cite{lit:Krahenbuhl_Koltun_2011} during the upsampling. We did not either include the two probes (cosine similarity-based k-means classification~\cite{lit:MacQueen1967} and supervised linear model with cross-entropy loss) as auxiliary loss-functions while training. In original STEGO implementation, the aforementioned steps, were also used to reduce the dimensionality of STEGO from its initial 70-dimensional output to the specific number of defined classes. In \ac{ac:RAPTOR} we include the dimensionality reduction in the final layer using only the \ac{ac:STEGO} loss function.

\subsection{Parcelization}
\label{sec:parcelization} 

The STEGO-model was originally designed as a self-supervised image segmentation algorithm. However, since it is self-supervised, it does not necessarily identify the exact clinically relevant anatomical segments. Instead, we have discovered that the automatically identified segments tend to be subsets of the clinically relevant cardiac substructures. Although Hamilton \textit{et at.}~\cite{lit:stego} do not specifically address this, it is also evident in their figures, where a single object is composed of several segments, such as a person's head or torso being part of different segmentation classes.

\begin{figure}[!h]
\centering
\includegraphics[width=0.48\columnwidth]{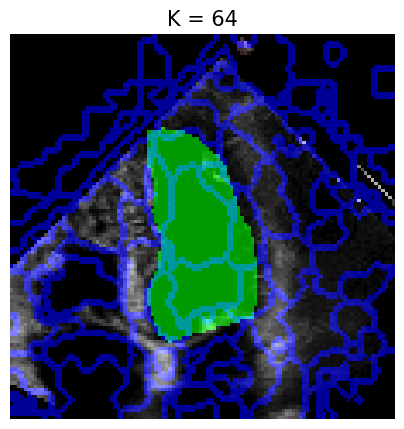}
\includegraphics[width=0.48\columnwidth]{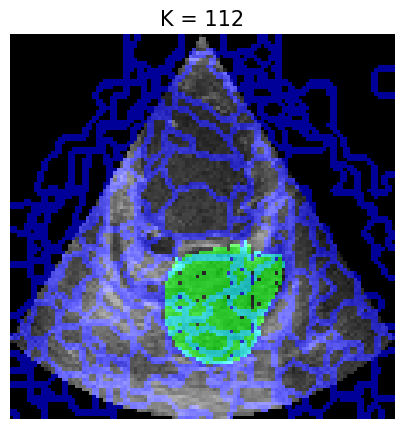}
\caption{Examples of parcelization of echocardiogram frames~\cite{lit:CAMUS,lit:EchoNetDynamic}. Blue lines denotes borders between parcels. The annotated segment is denoted in green.} 
\label{fig:parcelization_examples}
\end{figure}

To distinguish between these two concepts, we refer to the self-segmented output from the backbone as \textit{parcels} and the clinically relevant anatomical regions of interest as \textit{segments}. An example of how the parcels and a target segment relate to each other can be seen in Fig.~\ref{fig:parcelization_examples},
As a result, the backbone output (our \ac{ac:RAPTOR} combined with \ac{ac:DINO}) produces a \emph{parcelization} of the input image. 

The fact that the output of our completely self-supervised backbone consists of these interpretable and robustly identified (consistent in-between subjects) parcels is of great importance. That means that the backbone not only generates informative features for training smaller models using supervised techniques, but also produces features that are interpretable by humans. This interpretability is at the core of the ALAN-approach and sets it apart from general work on the concept of foundation models, pretraining, and self-supervision. The human interpretability of the backbone features makes it easier to design small white-box models for downstream tasks. The behavior of these white-box models can be easily communicated and parameterized. This also allow for expert elicitation, enabling the incorporation of domain knowledge into the algorithm, either by designing a mechanistic model or by defining a prior distribution on a probabilistic model. 

\subsection{Parcel-to-segment}
\label{sec:parcel-to-segment}

In the context of echocardiograms, a single parcel might not adequately delineate a desired anatomical region. Particularly, when dealing with a large number of segment classes, some parcels represent the same subregion in-between different sequences. In such scenarios, the union of the suitable subset of parcels effectively represents the target segment. 
In essence, achieving a reasonably accurate segmentation of the \ac{ac:LV} or \ac{ac:LA} can be as straightforward as uniting the parcels that encompass those regions. Therefore, segmenting a target region primarily involves identifying the parcels that overlap with it. 

We have designed a simple \emph{downstream} semantic segmentation model where the only unknown parameters are $K$ boolean variables that define which parcels should be joined to create targeted segment. 
Since the number of parameters is relatively small and easy to learn, the size of the training dataset does not need to be particularly large. 
The training process for this downstream model involves iterating through a validation subset for which ground truth labels are accessible. In the initial step, we identify parcels (based on their IDs) that overlap with ground truth annotations by at least 75\%. Next, we enforce that the overlap of over 75\% should be detected in a minimum of 50\% of the images in which these parcels are present. Additionally, these parcels must appear in at least 30\% of all images in the validation subset. Parcels consistently meeting these criteria are categorized as \emph{interior} to the target region.

The straightforward approach of segmenting by merging the \emph{interior} is functional but can be enhanced. 
Consequently, we incorporate prior knowledge that holds true for any cardiac cavity, such as the requirement for the segment to consist of a single connected component, absence of \textit{holes}, and the need for a smooth outline. 
We introduce this knowledge in a post-processing steps by: selecting the connected segment containing the most interior parcels (in cases where multiple disconnected segments are identified); considering exterior parcels whose neighboring borders consist solely of interior parcels as interior themselves on a per-image basis; and finally applying an active contour model~\cite{lit:Kass2004SnakesAC} to refine the segmentation edges.
We refer to this downstream model as \emph{parcel-to-segment}.

\section{Experiments}

\subsection{Datasets}
We evaluated the \ac{ac:ALAN}-approach for echocardiogram segmentation using one of the largest publicly available dataset -- EchoNet-Dynamic~\cite{lit:EchoNetDynamic} -- consisting of 10,030 \ac{ac:A4C} view echocardiogram sequences divided into training, validation, and test subsets. Each video is composed of a series of 112~$\times$~112~pixel grayscale images with segmentation of the left ventricle at the \ac{ac:ES} and \ac{ac:ED} frames. 
Additionally, we assessed the effectiveness of our methodology on the smaller \ac{ac:CAMUS}~\cite{lit:CAMUS} dataset. 
CAMUS consist of 450 publicly available \ac{ac:A4C} and \ac{ac:A2C} view sequences, with annotated segmentation masks of the \ac{ac:LV} endocardium, the \ac{ac:LV} epicardium, and the \ac{ac:LA} on selected \ac{ac:ES} and \ac{ac:ED} frames. 

The view classification was performed on \emph{Tufts Medical Echocardiogram Dataset 2} (TMED-2)~\cite{lit:huangSemisupervisedEchocardiogramBenchmark2021,lit:huangTMED2Dataset2022}. 
The dataset provides sets of labeled and unlabeled grayscale 112~x~112~pixel images of four different view types: parasternal long-axis (PLAX), parasternal short-axis (PSAX), apical two chamber (A2C), and apical four chamber (A4C). 
In our studies, we used 9~368 training- and 1~041 test-images that were distributed unequally across different view classes.

We exclusively used the EchoNet-Dynamic training subset to train the \emph{backbone}, whereas other subsets or datasets were employed for validation and testing. Fig.~\ref{fig:pipeline} displays the dataset splits required for each phase of the ALAN pipeline.

\begin{figure}[!h]
\centering
\includegraphics[width=\columnwidth]{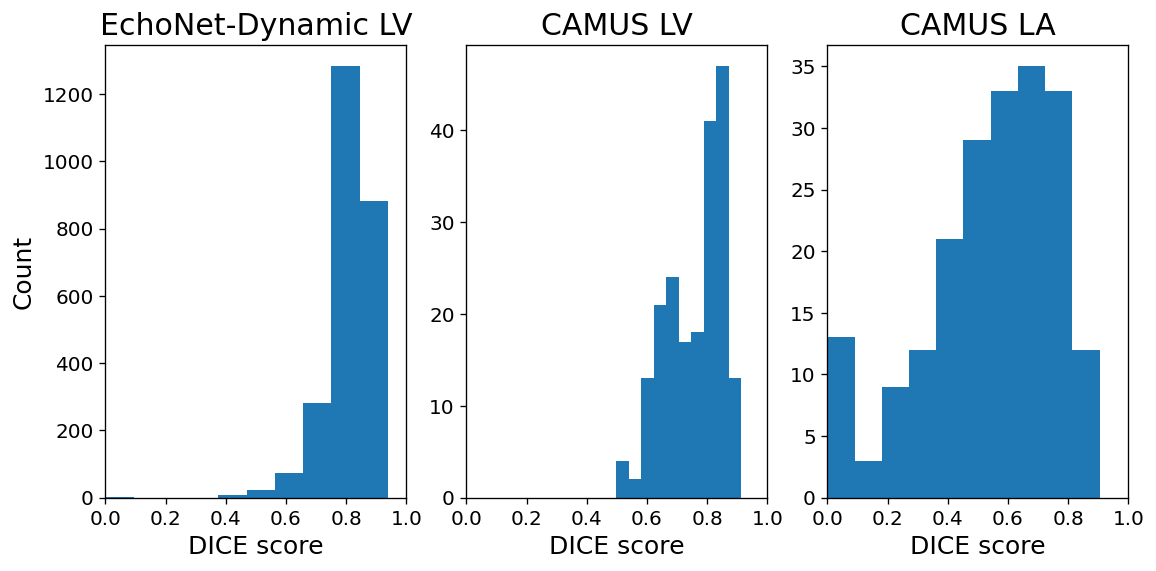}
\caption{Histograms of the calculated DICE scores for each end-systole and end-diastole frame across the test subsets of EchoNet-Dynamic (n=1277) and CAMUS (n=100) for \ac{ac:LV} and \ac{ac:LA} segmentation tasks. Prediction are made by DINO ViT-S/4 + \ac{ac:RAPTOR} trained on EchoNet-Dynamic with $K=64$.} 
\label{fig:DICE_hists}
\end{figure}

\begin{table*}[!h]
\centering
\caption{Segmentation results (DICE score) on test set of EchoNet-Dynamic~\cite{lit:EchoNetDynamic} and CAMUS~\cite{lit:CAMUS} datasets (accumulated over A2C and A4C views) for DINO backbone pretrained on 2 different training datasets. We report average $\pm$ standard deviation for \ac{ac:ES} and \ac{ac:ED} frame calculated across the test subsets of EchoNet-Dynamic (n=1277) and CAMUS (n=100). Parcelization was made using the STRAPTOR model with $K$ parcel classes. The best results for each anatomical region are highlighted in bold.}
\label{tab:all_par2seg_dice}
\resizebox{\textwidth}{!}{%
\begin{tabular}{cccccccc}
\toprule
\textbf{Training} & \multirow{2}{*}{\textbf{K}} & \multicolumn{2}{c}{\textbf{EchoNet-Dynamic}}                                   & \multicolumn{4}{c}{\textbf{CAMUS}}                                                                                                                        \\ \cmidrule(l){3-4} \cmidrule(l){5-8} 
\textbf{data} &  & \textbf{LV$_\mathbf{ES}$ {[}\%{]}}                       & \textbf{LV$_\mathbf{ED}$ {[}\%{]}}                       & \textbf{LV$_\mathbf{ES}$ {[}\%{]}}                        & \textbf{LV$_\mathbf{ED}$ {[}\%{]}}                       & \textbf{LA$_\mathbf{ES}$ {[}\%{]}}   & \textbf{LA$_\mathbf{ED}$ {[}\%{]}} \\ \midrule 
%                  & 8  &  $24.34 \pm 27.97$   &  $50.97 \pm 21.71$ &     $ 41.43\pm 26.10$   &  $ 57.77\pm 15.63$   &   $ 14.24\pm 13.12$ & $17.71 \pm 10.92$ \\
%                  & 16 &  $45.39 \pm 9.88$    &  $49.45 \pm 9.84$  &     $52.44 \pm 11.77$   &  $56.37 \pm 8.02$   &   $27.24 \pm 16.94$ & $26.13 \pm 14.04$ \\
\textbf{EchoNet}  & 32 & $71.91 \pm 10.41$    & $65.16 \pm 9.17$   &     $61.33 \pm 13.40$   &  $62.80 \pm 8.33$   &   $47.27 \pm 22.94$ & $45.94 \pm 19.63$ \\
\textbf{Dynamic}  & 64 & $\mathbf{80.11 \pm 8.53}$ & $\mathbf{81.81 \pm 6.58}$ & $73.85 \pm  9.51$ & $78.13 \pm 8.82$ & $54.50 \pm 23.48$ & $53.80 \pm 19.77$ \\
                  & 112& $ 76.44 \pm 8.88$    & $ 76.65\pm 7.64$   &     $\mathbf{75.67 \pm 9.50}$ &  $\mathbf{80.00 \pm 9.00}$   &  $\mathbf{55.49 \pm 18.28}$  &  $\mathbf{61.53 \pm 17.40}$ \\ \midrule   
\textbf{ImageNet} & 64 & $74.55 \pm 13.20$    & $79.09 \pm 9.45$   & $68.25 \pm 14.00$    & $74.89 \pm 9.09$    & $47.64 \pm 22.37$   & $53.16 \pm 21.33$  \\
%                  & 112 & $77.52 \pm 11.83$    & $78.47 \pm 8.94$   & $75.82 \pm 13.32$    & $77.90 \pm 9.34$    & $26.38 \pm 22.21$   & $52.46 \pm 25.00$  \\
\bottomrule
\end{tabular}%
}
\end{table*}

\begin{figure*}[!h]
     \centering
     \begin{subfigure}[b]{\textwidth}
         \centering
         \includegraphics[width=.2\textwidth]{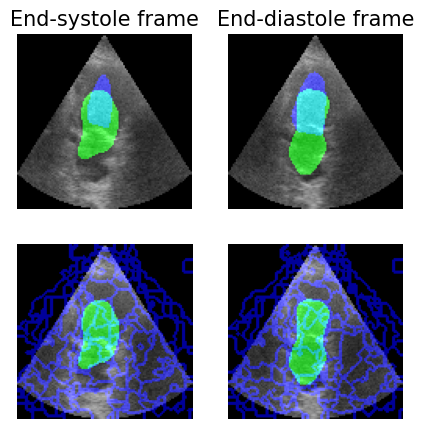}
         \includegraphics[width=.2\textwidth]{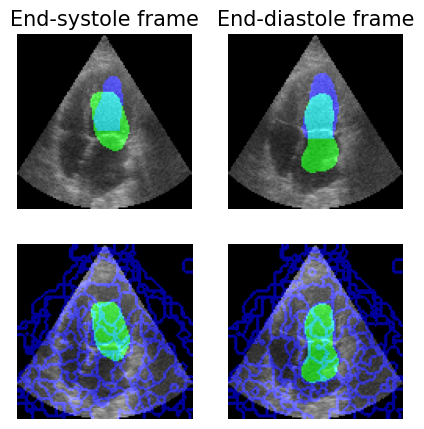}
         \includegraphics[width=.2\textwidth]{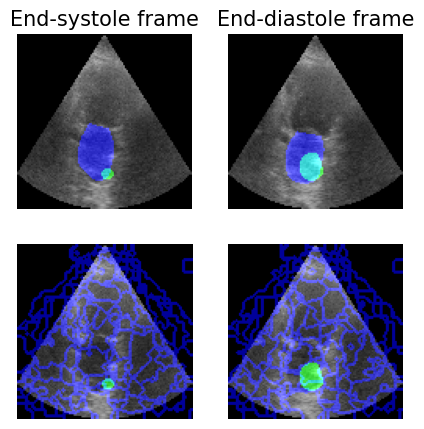}
         \includegraphics[width=.2\textwidth]{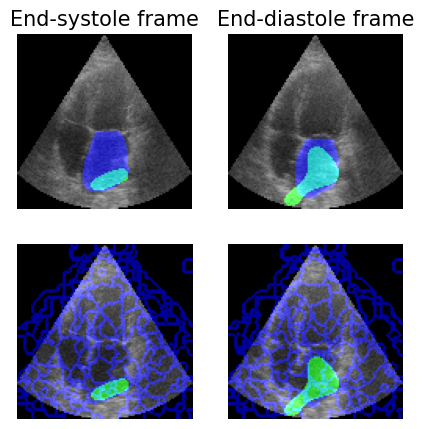}
         \caption{DICE $< 60\%$}
     \end{subfigure}
     \hfill
     \begin{subfigure}[b]{\textwidth}
         \centering
         \includegraphics[width=.2\textwidth]{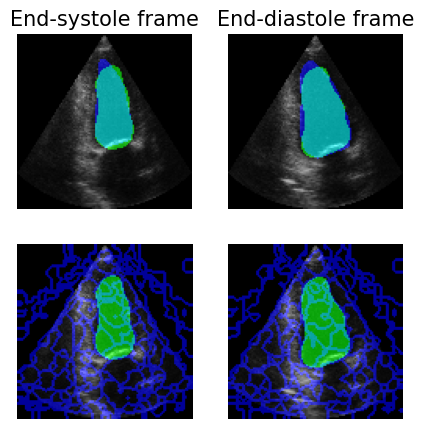}
         \includegraphics[width=.2\textwidth]{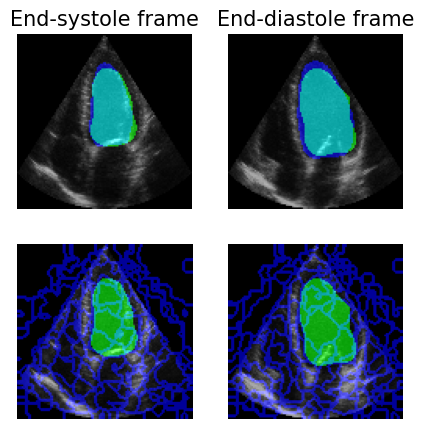}
         \includegraphics[width=.2\textwidth]{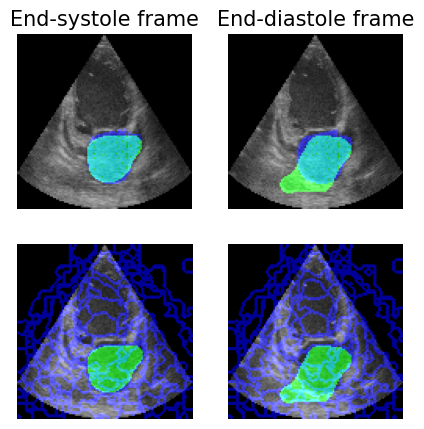}
         \includegraphics[width=.2\textwidth]{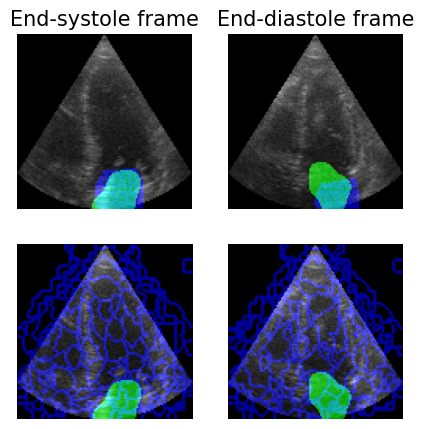}
         \caption{DICE $> 80\%$}
     \end{subfigure}
\caption{Parcel to segment visualizations for the selected CAMUS~\cite{lit:CAMUS} sequences with manual (blue) and predicted (green) masks (overlap in turquoise). Prediction are made by DINO ViT-S/4 + RAPTOR trained on EchoNet-Dynamic with $K=112$. The contours of all parcels are gathered in bottom rows.} 
\label{fig:more_examples_camus}
\end{figure*}

\begin{figure*}[!t]
     \centering
     \begin{subfigure}[b]{.45\textwidth}
         \centering
         \includegraphics[width=.45\textwidth]{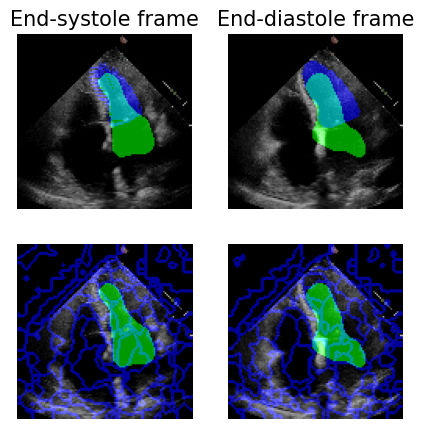}
         \includegraphics[width=.45\textwidth]{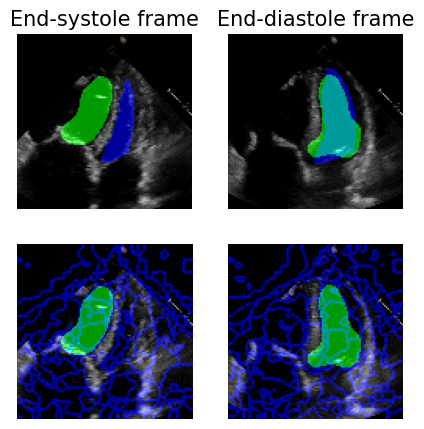}
         \caption{DICE $< 50\%$}
     \end{subfigure}
     \hfill
     \begin{subfigure}[b]{.45\textwidth}
         \centering
         \includegraphics[width=.45\textwidth]{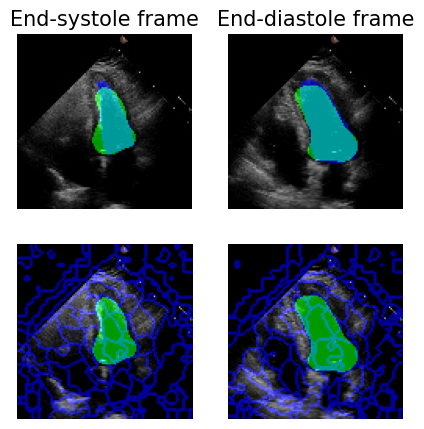}
         \includegraphics[width=.45\textwidth]{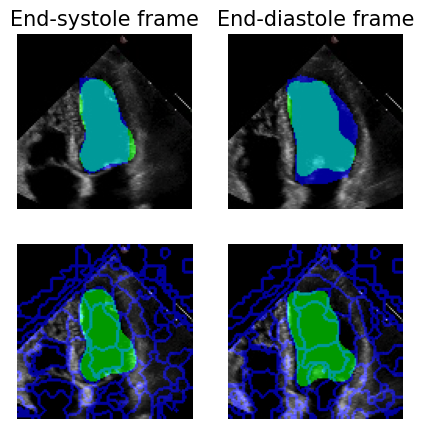}
         \caption{DICE $> 85\%$}
     \end{subfigure}
\caption{Parcel to segment visualizations for the selected EchoNet-Dynamic~\cite{lit:EchoNetDynamic} sequences with manual (blue) and predicted (green) masks (overlap in turquoise). Prediction are made by DINO ViT-S/4 + RAPTOR trained on EchoNet-Dynamic with $K=64$. The contours of all parcels are gathered in bottom row.} 
\label{fig:more_examples_echo}
\end{figure*}

\subsection{Echocardiographic image segmentation}

In the context of echocardiographic image segmentation, the \emph{backbone} comprises both the DINO and \ac{ac:RAPTOR} models working in conjunction. DINO undergoes independent training without the inclusion of the \ac{ac:RAPTOR} projection head. When training \ac{ac:RAPTOR}, the output of the frozen DINO model is used as input, i.e., DINO has by then already been trained. The dataset employed for training \ac{ac:RAPTOR} can either be the same as the one used for training DINO or a completely different dataset. Our \ac{ac:RAPTOR} is parcelizing the image into a small number of parcels, denoted as $K$, essentially performing the dimensionality reduction on the DINO output. Consequently, this parcelization process is more sensitive to the characteristics of the images it is trained on. This suggests that \ac{ac:RAPTOR} should be trained on echocardiograms, particularly in situations where the DINO model has been trained on an out-of-domain dataset like ImageNet. Consequently, the latent features from DINO that were provided as input for training the \ac{ac:RAPTOR} were all acquired using the training set of the EchoNet-Dynamic data. This holds true both for the DINO model trained on the very same dataset as well as the DINO model trained on ImageNet. For training \ac{ac:RAPTOR} we used 16 frames, consecutive but one frame apart, from each echo sequence. The start frame was chosen randomly in the sequences, as long as enough consecutive frames were available. In this way we ensured that at least a whole cardiac cycle was represented in each sequence and epoch~\cite{lit:EchoNetDynamic}, while still reducing the amount of data needed to be stored on the GPU at once. 

Considering that DINO's ViT-S architecture is the most complex model in the pipeline, we conducted an investigation to determine whether it is necessary for this model to be trained on actual echocardiogram images or if training it on a large, general dataset can capture essential latent features equally well. 
Therefore, we utilized the {DINO ViT-S/8}\footnote{\url{https://dl.fbaipublicfiles.com/dino/dino_deitsmall8_pretrain/dino_deitsmall8_pretrain.pth}} pretrained on ImageNet with an image size of 224~$\times$~224~pixels and a patch size of 8 (the spatial dimensional reduction in the first layer of the ViT).
Subsequently, we also pretrained a DINO ViT-S/4 on the EchoNet-Dynamic dataset. This involved using all individual frames from the training sequences, with each frame resized to 112~$\times$~112~pixels, and consequently set the DINO patch size to 4 to match the resolution and computational burden.

To carry out semantic segmentation in the downstream task we employed the parcel-to-segment method, as detailed in Section~\ref{sec:parcel-to-segment}. For this purpose we utilized the validation subset of the EchoNet-Dynamic dataset and the training subset of the CAMUS dataset. During the evaluation step the backbone was employed to extract the relevant parcels from the \ac{ac:ES} and \ac{ac:ED} frames of each test sequence, which were then used as input to the parcel-to-segment downstream model with fixed parameters, as illustrated in Fig.~\ref{fig:pipeline}.

Both the DINO backbone and the \ac{ac:RAPTOR} projection head are implemented in PyTorch. DINO was trained using the AdamW optimizer~\cite{lit:adamw}, with a linear ramp-up of the learning rate in the first 10 epochs, and a total training period of 40 epochs. The training parameters followed the original DINO schedule, except for modifications related to image size, patch size, and the deactivation of half precision. During the training of the \ac{ac:RAPTOR} projection head, we used the Adam optimizer with a learning rate of \mbox{$5e-3$}. It was observed that the network loss values reached saturation after approximately 40 epochs of training. Additionally, before training the \ac{ac:RAPTOR} projection head, we standardized the echocardiogram sequences based on the mean and standard deviation calculated for each image channel across the entire training subset. The post-processing step in the downstream model includes the following operations: identification of enclave parcels with a cutoff set to 8, spatial closing with radius set to 10, computing active contours~\cite{lit:Kass2004SnakesAC} with $\beta = 0.2$, $\gamma=0.01$, a maximum number of 10 iteration to optimize snake shape, and a parameter controlling attraction to edges set to 1.

Table~\ref{tab:all_par2seg_dice} summarizes results for EchoNet-Dynamic and CAMUS, testing subsets with respect to DICE score~\cite{lit:dice1945measures} to address the amount of overlap between the annotated \ac{ac:LV} and the predicted final segmentation. We observe that the size of the target region has a large impact on the performance of the segmentation, as the largest variability for calculated DICE scores is recorded for \ac{ac:LA} segmentation (see Fig.~\ref{fig:DICE_hists}). This can be due to a fewer number of parcels making up this region, the low spatial resolution of the DINO model due to patching, or the fact that manual annotations of the \ac{ac:LA} area sometimes does not cover the full \ac{ac:LA} surface~\cite{lit:CAMUS}. 
In general, the images with the worst DICE score usually exhibit lower quality (blurred, discoloured) and with the probe angle a bit off compared to the typical \ac{ac:A4C} and \ac{ac:A2C} views. The predicted segmentation mask in these cases often show cavities or extending out beyond the annotated edges in some direction(s). A deterioration in the predicted segmentation mask's integrity due to lower video quality was also observed in previous fully supervised approaches, where the average accuracy exceeded 85\% for both datasets (see Table~\ref{tab:sota_seg})~\cite{lit:Chen2021,lit:EchoNetDynamic,lit:saeed2022,lit:Stough2020LeftVA,lit:Wei2020}. Expert clinicians annotations can also vary in between them~\cite{lit:EchoNetDynamic}. Qualitative results of our studies are presented in Fig.~\ref{fig:more_examples_camus}--\ref{fig:more_examples_echo}.

\subsection{Per-frame view classification}

In order to classify the view types of echocardiograms we use a weighted kNN-classifier\cite{lit:wu2018unsupervised} on features extracted from our DINO ViT-S output~\cite{lit:DINO} (see Fig.~\ref{fig:pipeline}). 
We freeze the pretrained DINO ViT-S model and compute and store the features of the training subset of the TMED-2 dataset, for later \emph{downstream} view classification. To classify a test image, we compute its latent feature representation and compare it against all stored training features.
Similarly, as in the case of the echocardiogram segmentation we report the results both for the DINO ViT-S model trained on data from ImageNet and trained on data from EchoNet-Dynamic. For the kNN-classification we used the same parameters as in~\cite{lit:DINO}, except for the k-value itself. The classifiers performance in terms of averaged accuracy, as well as accuracy per class, for two selected $k$ values is shown in Table~\ref{tab:knn_acc}. 
\begin{table}[!hb]
    \centering
    \caption{Number of instances and reported accuracy of cardiac view classification on TMED-2 with k-nearest neighbor classification for DINO \ac{ac:ViT} trained on 2 different training datasets: ImageNet and EchoNet-Dynamic. The best results for each column are highlighted in bold.}
    \label{tab:knn_acc}
    \resizebox{\columnwidth}{!}{%
    \begin{tabular}{llcccccc}
    \toprule
     \textbf{Pretrain set} &  & \textbf{A2C} & \textbf{A4C}  & \textbf{PLAX} & \textbf{PSAX} & \textbf{Total} \\ \midrule 
     \multirow{2}{*}{\textbf{\# images}} & Train  & 1986 & 1504 & 4328   & 1554  & 9368  \\
     
      & Test   & 222  & 168  & 482    & 173   & 1041  \\ \midrule
    \textbf{EchoNet}                    & $k=2$  & \textbf{91.02} & \textbf{89.14}  &   95.01  &  80.81 & \textbf{90.78} \\
    \textbf{Dynamic}                    & $k=10$ & 87.43 & 87.33 & 95.01 & 79.07 & 89.53 \\ \midrule   
    \multirow{2}{*}{\textbf{ImageNet}}  & $k=2$  & 79.04 & 87.78 & 97.09 & 80.81 & 89.53 \\
                                        & $k=10$& 77.25 & \textbf{89.14} & \textbf{98.34} & \textbf{81.98} & 90.30 \\
    \bottomrule
    \end{tabular}%
    }
\end{table}
Our best approach is within a 7~percentage points margin of achieving the average accuracy reported by the semi-supervised method used by the creators of the TMED-2 dataset~\cite{lit:huangSemisupervisedEchocardiogramBenchmark2021,lit:huangTMED2Dataset2022}.

\section{Discussion and conclusions}

This paper presents a systematic study of a non-contrastive \ac{ac:SSL}-method as foundation for the \ac{ac:ALAN}-approach to echocardiogram analysis. ALAN distinguishes itself from previous self-supervised approaches by having interpretable latent features generated by the backbone, something that accommodates utilization of small and easily understood models for solving downstream tasks. This has the potential to streamline regulatory approval processes, improving medical comprehension, and enabling training on smaller datasets. Something that in turn can reduce costs of clinical trials and make inference on rare diseases possible.

We explore parcelization and manual engineering of parcel operations to segment crucial anatomical regions of the heart. Across different patients and datasets, specific sets of parcel classes consistently represent given cardiac regions. This noteworthy outcome is attributed to the combined use of \ac{ac:ViT} architecture and \ac{ac:DINO} training. In the RAPTOR-training, the number of parcel classes does not need to correspond to the actual segments in an image. Our findings indicate that a larger number of parcel classes is preferable, as it improves the segmentation of even small anatomical regions. However, the number of classes can still be kept below 100 to generate good results.
Furthermore, while our results for segmentation and view-classification indicate that self-supervised methods are approaching the efficiency of supervised methods, more work is still needed to enhance their performance in real-life medical applications.

The EchoNet-Dynamic dataset used for training exhibits lack of diversity, as it primarily consists only of apical chamber view classes. This may affect the ability to extrapolate our results to datasets with a greater variety of views available. However, obtained results for the DINO model trained on ImageNet proved that our \emph{backbone} can learn representations that capture rich data semantics and generalize across multiple image modalities. The work shows that a DINO model trained on general visual objects performs almost on-par with the corresponding model trained specifically on echocardiograms. In future studies we would like to investigate this further by training the \emph{backbone} model on datasets with other medical imaging modalities. 

\section{Acknowledgements}
This work has been carried out during the \textit{Eye for AI Program} thanks to the support of AstraZeneca and AI Sweden. Currently S.M. is a fellow of the AstraZeneca Postdoctoral Research Programme.


\begin{thebibliography}{10}\itemsep=-1pt

\bibitem{lit:anand2022}
Deepa Anand, Pavan Annangi, and Prasad Sudhakar.
\newblock Benchmarking self-supervised representation learning from a million
  cardiac ultrasound images.
\newblock In {\em 2022 44th Annual International Conference of the IEEE
  Engineering in Medicine \& Biology Society (EMBC)}, pages 529--532. IEEE,
  2022.

\bibitem{lit:Azizi9710396}
Shekoofeh Azizi, Basil Mustafa, Fiona Ryan, Zachary Beaver, Jan Freyberg,
  Jonathan Deaton, Aaron Loh, Alan Karthikesalingam, Simon Kornblith, Ting
  Chen, Vivek Natarajan, and Mohammad Norouzi.
\newblock Big self-supervised models advance medical image classification.
\newblock In {\em 2021 IEEE/CVF International Conference on Computer Vision
  (ICCV)}, pages 3458--3468, 2021.

\bibitem{lit:DINO}
Mathilde Caron, Hugo Touvron, Ishan Misra, Hervé Jégou, Julien Mairal, Piotr
  Bojanowski, and Armand Joulin.
\newblock Emerging properties in self-supervised vision transformers.
\newblock In {\em 2021 IEEE/CVF International Conference on Computer Vision
  (ICCV)}, pages 9630--9640, 10 2021.

\bibitem{lit:chaitanya2020contrastive}
Krishna Chaitanya, Ertunc Erdil, Neerav Karani, and Ender Konukoglu.
\newblock Contrastive learning of global and local features for medical image
  segmentation with limited annotations.
\newblock {\em Advances in Neural Information Processing Systems}, 33, 2020.

\bibitem{lit:chartsias2021}
Agisilaos Chartsias, Shan Gao, Angela Mumith, Jorge Oliveira, Kanwal Bhatia,
  Bernhard Kainz, and Arian Beqiri.
\newblock Contrastive learning for view classification of echocardiograms.
\newblock In J.~Alison Noble, Stephen Aylward, Alexander Grimwood, Zhe Min,
  Su-Lin Lee, and Yipeng Hu, editors, {\em Simplifying Medical Ultrasound},
  pages 149--158, Cham, 2021. Springer.

\bibitem{lit:ChenPSA17}
Liang{-}Chieh Chen, George Papandreou, Florian Schroff, and Hartwig Adam.
\newblock Rethinking atrous convolution for semantic image segmentation.
\newblock {\em CoRR}, abs/1706.05587, 2017.

\bibitem{lit:SimCLR}
Ting Chen, Simon Kornblith, Mohammad Norouzi, and Geoffrey~E. Hinton.
\newblock A simple framework for contrastive learning of visual
  representations.
\newblock {\em CoRR}, abs/2002.05709, 2020.

\bibitem{lit:Chen2021}
Yida Chen, Xiaoyan Zhang, Christopher~M Haggerty, and Joshua~V Stough.
\newblock Assessing the generalizability of temporally coherent
  echocardiography video segmentation.
\newblock In {\em Medical Imaging 2021: Image Processing}, volume 11596, pages
  463--469. SPIE, 2021.

\bibitem{lit:dice1945measures}
Lee~Raymond Dice.
\newblock Measures of the amount of ecologic association between species.
\newblock {\em Ecology}, 26(3):297--302, July 1945.

\bibitem{lit:vit}
Alexey Dosovitskiy, Lucas Beyer, Alexander Kolesnikov, Dirk Weissenborn,
  Xiaohua Zhai, Thomas Unterthiner, Mostafa Dehghani, Matthias Minderer, Georg
  Heigold, Sylvain Gelly, Jakob Uszkoreit, and Neil Houlsby.
\newblock An image is worth 16x16 words: Transformers for image recognition at
  scale.
\newblock In {\em Proceedings of the International Conference on Learning
  Representations}, 2021.

\bibitem{lit:GoodBengCour16}
Ian~J. Goodfellow, Yoshua Bengio, and Aaron Courville.
\newblock {\em Deep Learning}.
\newblock MIT Press, Cambridge, MA, USA, 2016.
\newblock \url{http://www.deeplearningbook.org}.

\bibitem{lit:byol}
Jean-Bastien Grill, Florian Strub, Florent Altch\'{e}, Corentin Tallec, Pierre
  Richemond, Elena Buchatskaya, Carl Doersch, Bernardo Avila~Pires, Zhaohan
  Guo, Mohammad Gheshlaghi~Azar, Bilal Piot, koray kavukcuoglu, Remi Munos, and
  Michal Valko.
\newblock Bootstrap your own latent - a new approach to self-supervised
  learning.
\newblock In H. Larochelle, M. Ranzato, R. Hadsell, M.F. Balcan, and H. Lin,
  editors, {\em Advances in Neural Information Processing Systems}, volume~33,
  pages 21271--21284. Curran Associates, Inc., 2020.

\bibitem{lit:stego}
Mark Hamilton, Zhoutong Zhang, Bharath Hariharan, Noah Snavely, and William~T
  Freeman.
\newblock Unsupervised semantic segmentation by distilling feature
  correspondences.
\newblock {\em CoRR}, abs/2203.08414, 2022.

\bibitem{lit:hendrycks2023gaussian}
Dan Hendrycks and Kevin Gimpel.
\newblock Gaussian error linear units (gelus), 2023.

\bibitem{lit:huangSemisupervisedEchocardiogramBenchmark2021}
Zhe Huang, Gary Long, Benjamin Wessler, and Michael~C. Hughes.
\newblock A new semi-supervised learning benchmark for classifying view and
  diagnosing aortic stenosis from echocardiograms.
\newblock In {\em Proceedings of the 6th Machine Learning for Healthcare
  Conference (MLHC)}, 2021.

\bibitem{lit:huangTMED2Dataset2022}
Zhe Huang, Gary Long, Benjamin Wessler, and Michael~C. Hughes.
\newblock Tmed 2: A dataset for semi-supervised classification of
  echocardiograms.
\newblock In {\em Technical Report}, 2022.

\bibitem{lit:Kass2004SnakesAC}
Michael Kass, Andrew~P. Witkin, and Demetri Terzopoulos.
\newblock Snakes: Active contour models.
\newblock {\em International Journal of Computer Vision}, 1:321--331, 2004.

\bibitem{lit:Krahenbuhl_Koltun_2011}
Philipp Krahenbuhl and Vladlen Koltun.
\newblock Efficient inference in fully connected crfs with gaussian edge
  potentials.
\newblock In {\em Proceedings of the Neural Information Processing Systems},
  2011.

\bibitem{lit:CAMUS}
Sarah Leclerc, Erik Smistad, Joao Pedrosa, Andreas Østvik, Frederic
  Cervenansky, Florian Espinosa, Torvald Espeland, Erik Berg, Pierre-Marc
  Jodoin, T. Grenier, Carole Lartizien, Jan D'hooge, Lasse Løvstakken, and
  Olivier Bernard.
\newblock Deep learning for segmentation using an open large-scale dataset in
  2d echocardiography.
\newblock {\em IEEE transactions on medical imaging}, 38(9):2198--2210, 08
  2019.

\bibitem{lit:Li23}
Zihan Li, Yunxiang Li, Qingde Li, Puyang Wang, Dazhou Guo, Le Lu, Dakai Jin,
  You Zhang, and Qingqi Hong.
\newblock Lvit: Language meets vision transformer in medical image
  segmentation.
\newblock {\em IEEE Transactions on Medical Imaging}, PP:12, 06 2023.

\bibitem{lit:adamw}
Ilya Loshchilov and Frank Hutter.
\newblock Fixing weight decay regularization in adam.
\newblock In {\em Proceedings of International Conference on Learning
  Representations}, 11 2018.

\bibitem{lit:MacQueen1967}
J.~B. MacQueen.
\newblock Some methods for classification and analysis of multivariate
  observations.
\newblock In L.~M.~Le Cam and J. Neyman, editors, {\em Proc. of the fifth
  Berkeley Symposium on Mathematical Statistics and Probability}, volume~1,
  pages 281--297. University of California Press, 1967.

\bibitem{lit:Nair2010RectifiedLU}
Vinod Nair and Geoffrey~E. Hinton.
\newblock Rectified linear units improve restricted boltzmann machines.
\newblock In {\em International Conference on Machine Learning}, 2010.

\bibitem{lit:EchoNetDynamic}
David Ouyang, Bryan He, Amirata Ghorbani, Neal Yuan, Joseph Ebinger, Paul
  A.~Heidenreich Curt P.~Langlotz, Robert~A. Harrington, David~H. Liang,
  Euan~A. Ashley, and James~Y. Zou.
\newblock Video-based ai for beat-to-beat assessment of cardiac function.
\newblock {\em Nature}, 580, 03 2020.

\bibitem{lit:Ronneberger2015}
Olaf Ronneberger, Philipp Fischer, and Thomas Brox.
\newblock U-net: Convolutional networks for biomedical image segmentation.
\newblock In Nassir Navab, Joachim Hornegger, William~M. Wells, and
  Alejandro~F. Frangi, editors, {\em MICCAI 2015}, pages 234--241, Cham, 2015.
  Springer.

\bibitem{lit:saeed2022}
Mohamed Saeed, Rand Muhtaseb, and Mohammad Yaqub.
\newblock Contrastive pretraining for echocardiography segmentation
  with limited data.
\newblock In Guang Yang, Angelica Aviles-Rivero, Michael Roberts, and
  Carola-Bibiane Sch{\"o}nlieb, editors, {\em MIUA}, pages 680--691, Cham,
  2022. Springer.

\bibitem{lit:Stough2020LeftVA}
Joshua~V. Stough, Sushravya Raghunath, Xiaoyan Zhang, John~M. Pfeifer,
  Brandon~K. Fornwalt, and Christopher~M. Haggerty.
\newblock Left ventricular and atrial segmentation of 2d echocardiography with
  convolutional neural networks.
\newblock In {\em Medical Imaging: Image Processing}, 2020.

\bibitem{lit:Hugo2021}
Hugo Touvron, Matthieu Cord, Matthijs Douze, Francisco Massa, Alexandre
  Sablayrolles, and Hervé Jégou.
\newblock Training data-efficient image transformers \& distillation through
  attention.
\newblock In {\em Proceedings of Machine Learning Research}, 2021.

\bibitem{lit:Wei2020}
Hongrong Wei, Heng Cao, Yiqin Cao, Yongjin Zhou, Wufeng Xue, Dong Ni, and Shuo
  Li.
\newblock Temporal-consistent segmentation of echocardiography with co-learning
  from appearance and shape.
\newblock In {\em Medical Image Computing and Computer Assisted
  Intervention--MICCAI 2020: 23rd International Conference, Lima, Peru, October
  4--8, 2020, Proceedings, Part II 23}, pages 623--632. Springer, 2020.

\bibitem{lit:wu2018unsupervised}
Zhirong Wu, Yuanjun Xiong, X~Yu Stella, and Dahua Lin.
\newblock Unsupervised feature learning via non-parametric instance
  discrimination.
\newblock In {\em Proceedings of the IEEE Conference on Computer Vision and
  Pattern Recognition}, 2018.

\bibitem{lit:ZHANG2021491}
Yuhan Zhang, Mingchao Li, Zexuan Ji, Wen Fan, Songtao Yuan, Qinghuai Liu, and
  Qiang Chen.
\newblock Twin self-supervision based semi-supervised learning (ts-ssl):
  Retinal anomaly classification in sd-oct images.
\newblock {\em Neurocomputing}, 462:491--505, 2021.

\bibitem{lit:Zhu2020Rubiks}
Jiuwen Zhu, Yuexiang Li, Yifan Hu, Kai Ma, S.~K. Zhou, and Yefeng Zheng.
\newblock Rubik's cube+: A self-supervised feature learning framework for 3d
  medical image analysis.
\newblock {\em Medical image analysis}, 64:101746, 2020.

\end{thebibliography}


{\small
        
}
\end{document}